\documentclass[10pt,twocolumn,letterpaper]{article}

\usepackage[pagenumbers]{iccv} 
\usepackage{multirow}
\usepackage{makecell}

\definecolor{iccvblue}{rgb}{0.21,0.49,0.74}
\usepackage[pagebackref,breaklinks,colorlinks,allcolors=iccvblue]{hyperref}

\def\paperID{6108} 
\def\confName{ICCV}
\def\confYear{2025}

\title{Grounded Chain-of-Thought for Multimodal Large Language Models}

\author{
Qiong Wu$^{1}$, Xiangcong Yang$^{1}$, Yiyi Zhou$^{1}$, Chenxin Fang$^{1}$, Baiyang Song$^{1}$, Xiaoshuai Sun$^{1}$, Rongrong Ji$^{1}$ \\
$^{1}$ Key Laboratory of Multimedia Trusted Perception and Efficient Computing, \\
Ministry of Education of China, Xiamen University, 361005, P.R. China. \\
{\tt\small \{qiong, yangxiangcong\}@stu.xmu.edu.cn, zhouyiyi@xmu.edu.cn,} \\ 
{\tt\small  fangchenxin@stu.xmu.edu.cn, songbaiyang07@gmail.com,  \{xssun, rrji\}@xmu.edu.cn}
}

\definecolor{myred}{rgb}{1.0, 0.0, 0.0}
\definecolor{mygreen}{rgb}{0.0, 0.69, 0.31}
\definecolor{myblue}{rgb}{0.0, 0.69, 0.94}
\definecolor{myyellow}{rgb}{1.0, 0.75, 0.0}

\begin{document}
\maketitle

\begin{abstract}
Despite great progress, existing \emph{multimodal large language models} (MLLMs) are prone to \emph{visual hallucination}, greatly impeding their trustworthy applications.
In this paper, we study this problem from the perspective of visual-spatial reasoning, and propose a new learning task for MLLMs, termed \emph{Grounded Chain-of-Thought} (\emph{\textbf{GCoT}}). 
Different from recent visual CoT studies, which focus more on visual knowledge reasoning, GCoT is keen to helping MLLMs to recognize and ground the relevant visual cues step by step, thereby predicting the correct answer with grounding coordinates as the intuitive basis. 
To facilitate this task, we also carefully design and construct a dataset called \emph{multimodal grounded chain-of-thought} (\emph{\textbf{MM-GCoT}}) consisting of 24,022 GCoT examples for 5,033 images. 
Besides, a comprehensive consistency evaluation system is also introduced, including  the metrics of \emph{answer accuracy} , \emph{grounding accuracy} and \emph{answer-grounding consistency}. 
We further design and conduct a bunch of experiments on 12 advanced MLLMs, and reveal some notable findings: 
\emph{i.} most MLLMs performs poorly on the consistency evaluation, indicating obvious visual hallucination; 
\emph{ii.}, visual hallucination is not directly related to the parameter size and general multimodal performance, \emph{i.e.}, a larger and stronger MLLM is not less affected by this issue. 
Lastly, we also demonstrate that the proposed dataset can help existing MLLMs to well cultivate their GCoT capability and reduce the inconsistent answering significantly. 
Moreover, their GCoT can be also generalized to exiting multimodal tasks, such as open-world QA and REC. 
Our dataset and evaluation scripts are anonymously released at: \url{https://github.com/DoubtedSteam/MM-GCoT}
\end{abstract}
%

\section{Introduction}
\label{sec:intro}

\begin{figure}
\centering
\includegraphics[width=1.0\columnwidth]{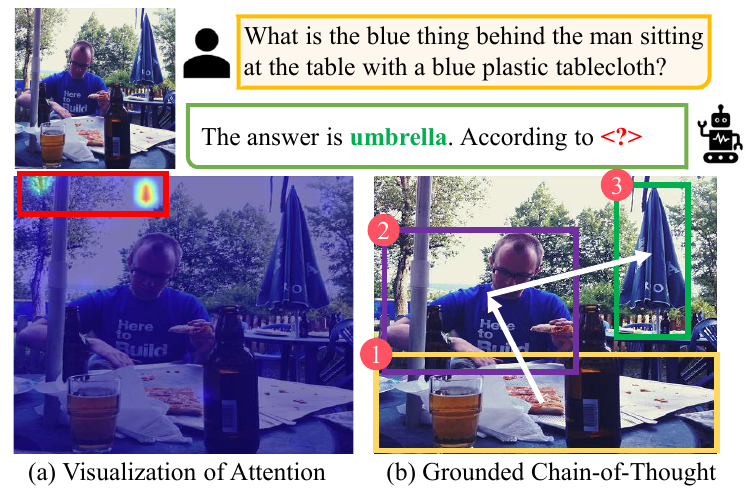}
\vspace{-5mm}
\caption{
The comparison between common question answering (a) and the proposed \emph{grounded chain-of-thought} (GCoT) (b). 
(a) An MLLM can directly output the correct answer, which however is often not based on what it, as shown in its attention heat map. 
This issue undermines its trustworthy application. 
(b) GCoT aims to help the MLLM make grounded reasoning step-by-step, and outputs the answer with coordinates as the intuitive basis. 
}
\label{Fig_motivation}
\vspace{-5mm}
\end{figure}

As a research hot-spot, \emph{Multimodal Large Language Model} (MLLM) \cite{llava-1.5, llava_ov, qwen-vl, qwen2.5-VL, Internvl, internvl2.5} has made remarkable progresses in approaching near-human vision-langauge (VL) capabilities on various benchmarks \cite{vqav2, mme, sqa-i, gqa}.
However, visual hallucination \cite{sun2024explore, zhuang2025vasparse} still remains a critical problem that long plagues the application of MLLMs. 
In practice, MLLMs may fail to answer the question due to insufficient understanding of the given image. 
This visual shortcoming is easy to be measured via existing  evaluation systems \cite{pope, hallucination, hallusionbench}, and it also arouses great attention and receive numerous recent efforts \cite{liu2023mitigating, zhuang2024game, wang2025mirage, zhuang2025vasparse, huo2024self}.

\begin{figure*}
\centering
\includegraphics[width=1.0\textwidth]{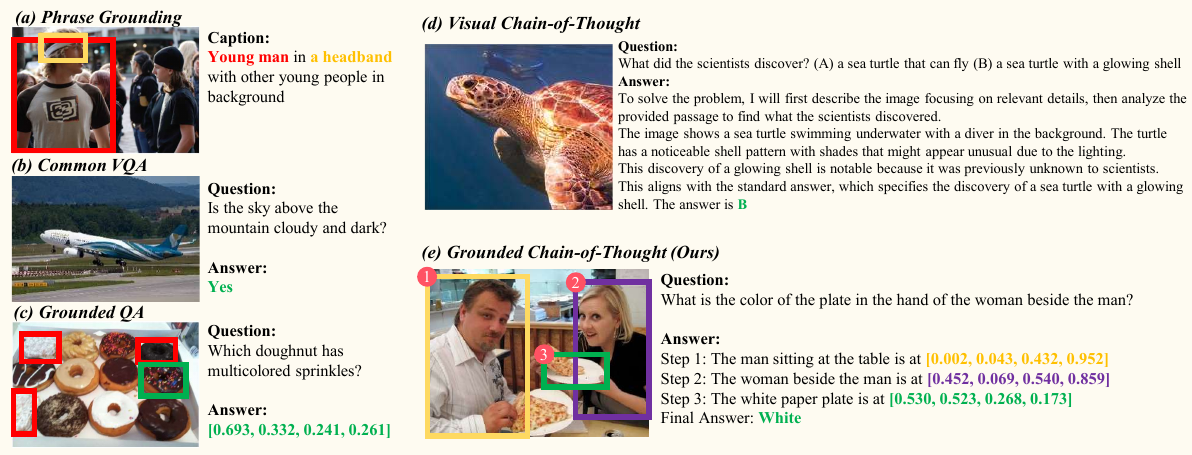}
\vspace{-7mm}
\caption{
Comparison between relevant vision-langauge tasks (a-d) and our GCoT (e). 
Phrase grounding (a) is similar to our GCoT in terms of grounding outputs, but it is only to detect mentioned instances in the caption. 
Grounded QA (c) also lacks the hidden reasoning steps compared to GCoT. 
VCoT (d) extends the common VQA of MLLMs (b) via providing more detailed answering thoughts, which is more about knowledge reasoning. 
In contrast, our GCoT aims to decompose the question into multiple task steps with grounded information, providing intuitive basis for the visual-spatial reasoning of MLLMs. 
}
\label{Fig_framework}
\vspace{-2mm}
\end{figure*}

A more subtle manifestation is when an MLLM can answer the question accurately, but relies more on data distribution bias than key visual information, also known as \emph{language bias} in previous VL study \cite{xu2024pride, zhou2024unibias, gupta2023bias}. 
As shown in Fig.\ref{Fig_motivation}, although the answer is correct, the MLLM attend to irrelevant regions for answering, so it also naturally ground the wrong answer cue. 
Although this issue does not affect the performance evaluation of MLLMs on most VL benchmarks, it is easy to misvalue their reliability and bring potential application risks. 
In this case, some efforts \cite{seqground, visual7w, plummer2015flickr30k} propose to force the model to ground the related instance in the image when answering, \emph{e.g.}, grounded QA \cite{visual7w}, or give more detailed explanations for their answers \cite{sqa-i, llava-cot}.

In this paper, we propose to alleviate the visual hallucination of MLLMs via enhancing their visual-spatial reasoning capability, and introduce a new learning task called \emph{grounded chain-of-thought} (GCoT), as shown in Fig.\ref{Fig_motivation}-(b).
Similar to the CoT research for LLMs \cite{kim2023cot, weifinetuned, si2023empirical}, GCoT is also to help the model make step-by-step reasoning before answering the question. 
But different from recent visual CoT works \cite{llava-cot, mme-cot, visualCoT, sqa-i}, which mainly focus on visual knowledge reasoning, GCoT aims to cultivate both visual and spatial perception capabilities of MLLMs, allowing them to ground the key steps for correctly answer the question. 
This property also make GCoT distinct from some relevant VL tasks, such as grounded QA \cite{visual7w} and visual grounding \cite{RefCOCO, seqground}, as shown in Fig.\ref{Fig_framework}.

\begin{figure*}
\centering
\includegraphics[width=1.0\textwidth]{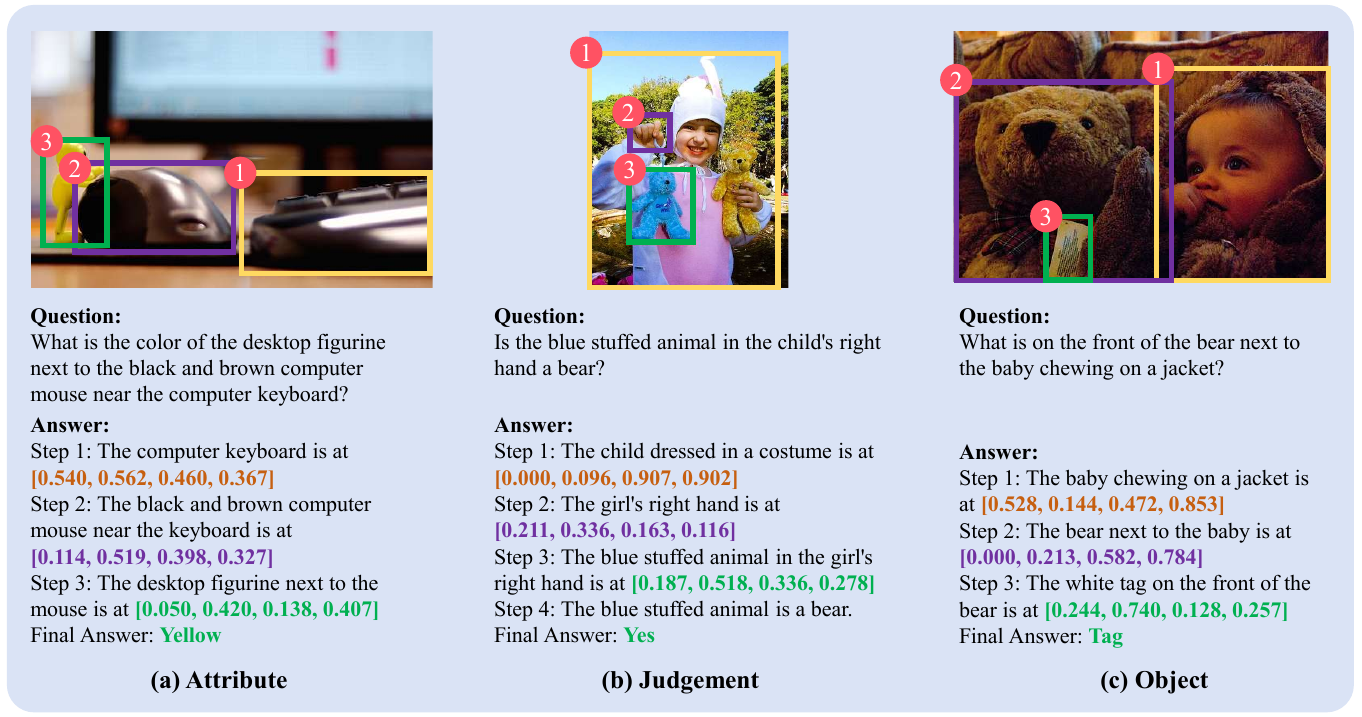}
\vspace{-7mm}
\caption{
Examples of the proposed \emph{multi-modal grounded chain-of-thought} (MM-GCoT). 
MM-GCoT has three splits of examples, namely \emph{Attribute} (a), \emph{Judgement} (b) and \emph{Object} (c). 
Each example consists of multiple reasoning steps with grounded information, \emph{i.e.}, the spatial coordinates, serving the cultivation of GCoT capability for MLLMs.  
Meanwhile, MM-GCoT also reserves a set of example for the hallucination evaluation of MLLMs in terms the metrics of \emph{answer accuracy}, \emph{grounding accuracy} and \emph{answer-grounding consistency}.
}
\label{Fig_datasample}
\vspace{-2mm}
\end{figure*}

To facilitate GCoT, we also propose a large dataset for the SFT tuning of existing MLLMs, named by \emph{multimodal and grounded chain-of-thought} (MM-GCoT).
MM-GCot consists of 23,028 training examples of three tasks, \emph{i.e.}, \emph{Attribute}, \emph{Judgment}, and \emph{Object}.
As shown in Fig.\ref{Fig_datasample}, an MM-GCoT example has multiple steps with grounding information. 
In this case, MLLMs can learn to reason key elements in the image for answering, and provide their coordinates as the intuitive basis, yielding a more trustworthy reasoning process.
Moreover, GCoT also reserves about 994 examples for the evaluation of the visual-spatial reasoning capability and visual hallucination of MLLMs, which is conducted by three metrics, namely \emph{answer accuracy}, \emph{grounding accuracy} and \emph{answer-grounding consistency}, respectively.

In the experimental section, we first evaluate a set of advanced MLLMs on our MM-GCoT benchmark, including LLaVA-1.5\cite{llava-1.5}, LLaVA-OneVision\cite{llava_ov}, Qwen2.5-VL\cite{qwen2.5-VL} and InternVL2.5\cite{internvl2.5}.
From these results, we summarize three notable findings about these MLLMs. 
Firstly, most MLLMs exhibit low answer-grounding consistency, indicating the great difference between their perception and reasoning. 
Thus, it is hard to ensure that their answers are what they see.  
Secondly, this consistency is not related to their grounding or reasoning capability, \emph{i.e.}, an MLLM may performs well on a single task of visual grounding or question answering, but is hard to maintain a good consistency on MM-GCoT. 
Last but not least, although super-large MLLMs has much stronger multimodal QA accuracies on commong benchmarks, but perform very poorly on MM-GCoT, showing obvious data over-fitting during testing. 
For instance, Qwen-VL-72B shows much worse answer-grounding consistency than its 7B version. 
These findings imply that visual hallucination related to data overfitting is prominent in existing MLLMs, especially the super-large ones. 
Afterwards, we further examine the benefit of GCoT by training MLLMs with our MM-GCoT data. 
The result show that MLLMs enhanced by GCoT capabilities can exhibit much better visual-spatial reasoning while reducing the inconsistency between the answer and grounded proofs, which also show the great potential of GCoT in alleviating visual hallucination.

Overall, our contributions are three-fold:
\begin{itemize}
\item We propose a new learning task to alleviate visual hallucination of MLLMs from the perspective of visual-spatial reasoning, termed \emph{grounded chain-of-thought} (GCoT).
\item We construct a new dataset called MM-GCoT to facilitate the research of GCoT for MLLMs, which also contains a benchmark to evaluate the hallucination of MLLMs.
\item The extensive experiments indicate that notable findings of existing MLLMs in terms of visual hallucination, and also validate the merits of GCoT capability for MLLMs.
\end{itemize}

\section{Related Work}

Recent advances in multimodal large language models (MLLMs) \cite{llava-1.5, llava_ov, qwen-vl, qwen2.5-VL, Internvl, internvl2.5} have demonstrated remarkable capabilities in vision-language tasks, significantly outperforming traditional vision-language models \cite{vilt, meter, blip2}. 
Despite their impressive performance, MLLMs still suffer from the long-lasting issue of visual hallucination.
In traditional vision-language research, visual hallucination primarily denote \emph{language bias} \cite{wu2025lanp, li2024naturalbench, han2024instinctive, zhang2024seeing}, where models tend to predict answers based on question-answer distribution bias rather than the observed visual information. 
Recent research mainly regards the visual hallucination of MLLMs as the problem of visual shortcoming \cite{liu2024paying, fu2024mitigating, qu2024look}, while the issue of language bias receives less attention. 
Several benchmark datasets have been developed to evaluate visual hallucination in MLLMs. 
For instance, POPE \cite{pope} introduces object-presence verification tasks.
While HallusionBench \cite{hallusionbench} through professionally designed question pairs and control groups to assess MLLMs' hallucination tendencies.
However, these benchmarks often rely on simple binary verification metrics, which is hard to judge the reliability of MLLMs' reasoning process. 
To this end, we focus more on examining the hallucination of MLLMs in a granular manner, especially about language prior.

In terms of \emph{chain-of-thought} (CoT) research for MLLMs, recent efforts \cite{llava-cot, sqa-i, visualCoT, mathvista} mainly focus on visual explanations or knowledge reasoning.
For example, LLaVA-CoT \cite{llava-cot} requires an MLLM to output the detailed solution about the problem, and then output the answer.
ScienceQA \cite{sqa-i} anchors its scientific reasoning in knowledge derived from pre-trained memory, where multimodal contexts activate latent domain information.
However, these approaches do not verify whether each reasoning step is correct visual evidence.

Our work is also closely related to visual grounding tasks \cite{RefCOCO, RefCOCOg, seqground, vg, visual7w, plummer2015flickr30k}.
Previous visual grounding task like \emph{phrase grounding} aim to locate all objects mentioned in a given caption \cite{RefCOCO, RefCOCOg}, our GCoT is to analyze the question and form the grounded reasoning steps, and then give the correct answer. 
\emph{Grounded Question Answering} (GroundedQA) is a extended visual grounding task, which also aims to verify the credibility of MLLM's outputs by providing the grounding information. However, GroundedQA lacks of the annotated reasoning steps to improve the model's visual-spatial capability \cite{visual7w, vqa-x}.
In contrast, our MM-CoT can help the models to identify and ground visual evidence throughout the entire reasoning process.
Another recent work similar to ours is VoCoT \cite{journals/corr/abs-2405-16919}, which also focuses on the grounded reasoning steps for MLLMs. 
In addition to the difference of the way to construct grounding datasets, our work also differs in the focus of alleviating visual hallucination in MLLMs, and proposes a set of new evaluation metrics for this issue.

The proposed GCoT can be also contributed to the research of \emph{Vision-Language-Action} (VLA) models for emboided applications \cite{yang2024thinking}.
By incorporating grounded spatial reasoning into the chain-of-thought process, our approach enables more accurate and interpretable decision-making in complex embodied tasks. 
This integration is particularly valuable for robotic navigation and manipulation scenarios where precise spatial understanding and step-by-step action planning are essential for successful task completion.

\section{Grounded Chain-of-Thought}

In this paper, we propose a new learning task for \emph{multimodal large langauge models} (MLLMs) termed \emph{grounded chain-of-thought} (GCoT). 
GCoT aims to alleviliate the visual hallucination issue of MLLMs via enhancing their grounded visual-spatial reasoning capability.
As shown in Fig.\ref{Fig_framework}, given a question an an image, GCoT will first let MLLMs to analyze and decompose the task, then reason about each task step and provide the spatial information of task-related elements in the image.  
Based on these grounded reasoning steps, the MLLM will predict the answer and provide the spatial coordinates as the intuitive basis.

\begin{table}[t]
\caption{
Statistics of each split of MM-GCoT
}
\vspace{-3mm}
\resizebox{1.0\columnwidth}{!}{
\begin{tabular}{l|rrr|c}
\toprule[1.00pt]
\multicolumn{1}{c|}{\textbf{Split}} & \multicolumn{1}{c}{\textbf{Attribute}} & \multicolumn{1}{c}{\textbf{Judgement}} & \multicolumn{1}{c|}{\textbf{Object}} & \textbf{Total} \\ 
\midrule
Trainval    & 5,183   & 12,849    & 4,996 & 23,028\\
Test        & 459     & 206       & 329   & 994 \\
\bottomrule[1.00pt]
\end{tabular}
\label{detail_MM_GCoT}
}
\vspace{2mm}

\caption{
Comparison between MM-GCoT and VL datasets. 
}
\vspace{-3mm}
\resizebox{1.0\columnwidth}{!}{
\begin{tabular}{l|rrcc}
\toprule[1.00pt]
\multicolumn{1}{c|}{\textbf{Datasets}} & \multicolumn{1}{c}{\textbf{TrainVal}} & \multicolumn{1}{c}{\textbf{Test}} & \multicolumn{1}{c}{\textbf{CoT}} & \textbf{Grounding} \\ 
\midrule
Visual7W \cite{visual7w}   & 229,557   & 98,382    &  & \checkmark \\
ScienceQA \cite{sqa-i}     & 16,967    & 4,241     & \checkmark & \\
MME-CoT \cite{mme-cot}     & -         & 1,130     & \checkmark & \\
\midrule
MM-GCoT       & 23,028    & 994     & \checkmark & \checkmark \\
\bottomrule[1.00pt]
\end{tabular}
\label{comparison_dataset}
}
\vspace{-2mm}
\end{table}

In this context, we can see that GCoT transform the direct answering of MLLMs, originally formulated as a simple mapping function $\mathcal{F}: \mathcal{I},\mathcal{T} \rightarrow \mathcal{A}$, into a multi-step decision process:
\begin{equation}
\begin{aligned}
P(\mathcal{A}|\mathcal{I},\mathcal{T})=\prod_{t=1}^T P(R_t, G_t|\mathcal{I},\mathcal{T},\mathbf{G}_{<t},\mathbf{R}_{<t}),
\end{aligned}
\label{objective}
\end{equation}
where $\mathcal{I}$ and $\mathcal{T}$ are input image and text.
$\mathcal{A}$ represents the final answer.
$\mathbf{G}_{t}$ and $\mathbf{R}_{t}$ denote the visual evidence and the textual reasoning state at step $t$.
From Eq.\ref{objective}, we can see the cultivation of GCoT can be accomplished by reinforcement-learning based schemes in addition to the SFT tuning used in this paper, \emph{e.g.}, GRPO \cite{shao2024deepseekmath} used in DeepSeek-R1 \cite{guo2025deepseek}. 
Thus, this setting provides ample room for practitioners to explore RL training on non-GCoT labeled data. 
Besides, its reasoning step is about the thinking of visual scenes with grounded spatial information, which is also different from existing (visual) CoT tasks.
In addition to MLLMs, this setting is also very critical for the exploration of \emph{vision-language-action} (VLA) models in embodied scenarios \cite{yang2024thinking}.


\begin{figure}
\centering
\includegraphics[width=1.0\columnwidth]{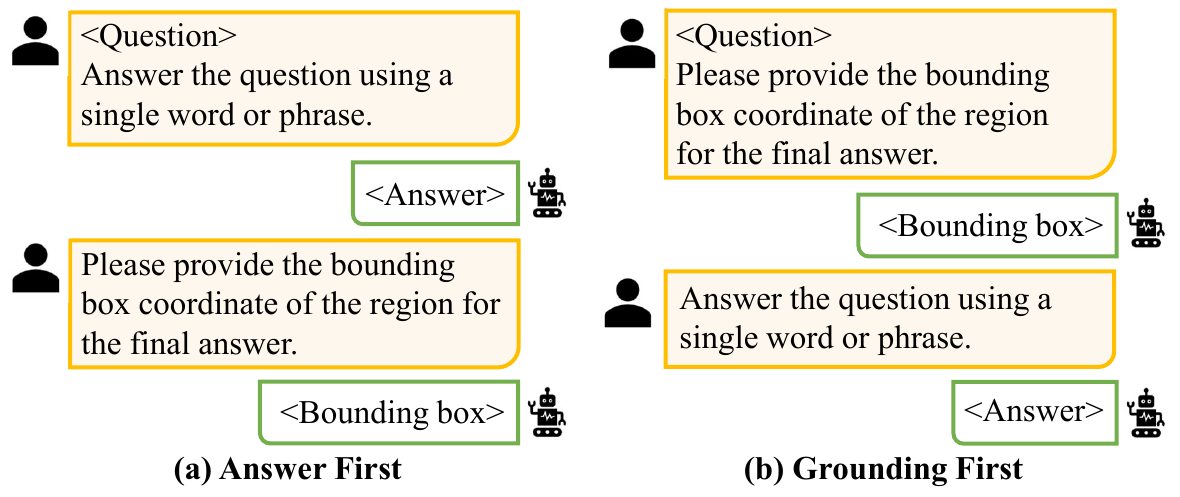}
\vspace{-7mm}
\caption{
Illustration of two prompting settings for evaluating LLaVA \cite{llava-1.5} on MM-GCoT.
(a) Answer-First: The MLLM generates a textual response, followed by producing the corresponding bounding box in subsequent conversational turns.
(b) Grounding-First: The MLLM initially provides a visual bounding box, then responses with a text answer.
}
\label{Fig_response}
\vspace{-3mm}
\end{figure}

\section{MM-GCoT: A New Multimodal Dataset}

To promote the research of GCoT, we also carefully design and construct a large SFT dataset for MLLMs, termed MM-GCoT. 
As shown in Tab.\ref{comparison_dataset}, our MM-GCoT is the only dataset that combines grounding with CoT.
Similar to \cite{llava-cot, visualCoT}, MM-GCoT can be used to train the basic GCoT capability of MLLMs, thereby facilitating the future exploration of RL or weakly-supervised training on more VL data. 
Besides, MM-GCoT also reserves a part of examples for the visual hallucination of MLLMs.

\noindent \textbf{Dataset Statistics.} MM-GCoT contains 23,028 training/validation samples for training GCoT and 994 test samples for the hallucination evaluation of MLLMs.
As shown in Fig.\ref{detail_MM_GCoT}, all samples are categorized into three types, namely \emph{Attribute}, \emph{Judgement} and \emph{Object}. 
Here, Attribute refers to questions provide the object's type and location while inquiring about specific attributes of the target.
Judgment is to examine the correctness of the given description, 
while Object is the question that focus on identifying the object category at a specified location.
%
%
\begin{table*}[t]
\caption{
Evaluation results of existing MLLMs on our MM-GCoT benchmark under the \emph{answer-first} prompting setting.
``A-Acc'' denotes the answer accuracy.
``G-Acc'' denotes the grounding accuracy.
``Consist.'' denotes the answer-grounding consistency.
%
%
The best and second best results are marked in \textbf{bold} and \underline{underline}, receptively.
}
\vspace{-3mm}
\resizebox{1.0\textwidth}{!}{
\begin{tabular}{l|ccc|ccc|ccc|ccc}
\toprule[1.00pt]
\multicolumn{1}{c|}{\multirow{2}{*}{\textbf{Method}}} & 
\multicolumn{3}{c|}{\textbf{Attribute}} & 
\multicolumn{3}{c|}{\textbf{Judgement}} & 
\multicolumn{3}{c|}{\textbf{Object}}    & 
\multicolumn{3}{c}{\textbf{Average}}  \\
\multicolumn{1}{c|}{}   & 
\textbf{A-Acc.}  & \textbf{G-Acc} & \textbf{Consist.} & 
\textbf{A-Acc.}  & \textbf{G-Acc} & \textbf{Consist.} & 
\textbf{A-Acc.}  & \textbf{G-Acc} & \textbf{Consist.} & 
\textbf{A-Acc.}  & \textbf{G-Acc} & \textbf{Consist.} \\
\midrule
LLaVA-7B               \cite{llava-1.5} & 68.6 & 9.2 & 8.8 & 83.0 & 11.2 & 11.5 & 58.4 & 9.1 & 9.9 & 70.0 & 9.8 & 10.1        \\
LLaVA-13B              \cite{llava-1.5} & 69.7 & 12.0 &9.6  & 83.5   & 14.6 & 15.4 & 58.7   & 11.6        & 14.4 & 70.6 & 12.7 & 13.1        \\
LLaVA-OneVision-0.5B   \cite{llava_ov} & 65.6 & 0.0 & 0.0  & 83.0   & 0.0  & 0.0  & 43.5   & 0.0         & 0.0  & 64.0 & 0.0  & 0.0         \\
LLaVA-OneVision-7B     \cite{llava_ov} & 69.8 & 11.2& 10.9  & \bf{89.8} & 12.1 & 11.7 & \bf{65.2}   & 10.5         & 10.7 & \underline{74.9} & 11.3 & 11.1        \\
LLaVA-OneVision-72B    \cite{llava_ov} & \underline{74.3}    & 13.5 & 11.0 & 88.3   & 17.0 & 17.9 & \underline{64.4}   & 19.1        & 18.5 & \bf{75.7} & 16.5 & 15.8        \\
InternVL2.5-4B         \cite{internvl2.5} & 63.4 & 21.8& 18.8 & 84.5   & 24.3 & 20.4 & 46.2 & 23.1        & 20.0 & 64.7 & 23.1 & 19.8        \\
InternVL2.5-8B         \cite{internvl2.5} & 61.9 & 51.9& 42.6 & \underline{89.3}   & 36.4 & 31.5 & 48.3        & 43.2        & 39.4 & 66.5    & 43.8 & 37.8        \\
InternVL2.5-38B        \cite{internvl2.5} & 63.2 & 53.2 & 44.3 & 84.0   & \underline{45.6} & \underline{43.5} & 53.8   & 42.9        & 38.9 & 67.0 & 47.2 & 42.2        \\
InternVL2.5-78B        \cite{internvl2.5} & 57.9 & 41.8 & 33.1 & 86.9   & 25.7 & 26.1 & 50.8   & \underline{52.9}        & 47.0 & 65.2 & 40.1 & 35.4        \\
Qwen2.5-VL-3B-Instruct \cite{qwen2.5-VL} & 71.5     & \underline{57.3}       & \underline{50.0} & 86.9 & 39.3    & 39.8        & 57.1 & 48.9    & \underline{51.7}        & 71.8 & \underline{48.5}    & \underline{47.2}        \\
Qwen2.5-VL-7B-Instruct \cite{qwen2.5-VL} & 73.6     & \bf{72.5}       & \bf{59.8} & 87.9 & \bf{56.3}    & \bf{51.5}        & 57.8 & \bf{64.1}    & \textbf{59.1}       & 73.1 & \bf{64.3}    & \bf{56.8}        \\
Qwen2.5-VL-72B-Instruct\cite{qwen2.5-VL} & \bf{75.5}     & 44.7       & 41.6 & 85.9 & 31.1    & 32.4        & 62.9 & 43.2    & 43.0        & 74.8 & 39.6    & 39.0        \\
\bottomrule[1.00pt]
\end{tabular}
}
\label{answer_first}
\vspace{2mm}

\caption{
Evaluation results of existing MLLMs on our MM-GCoT benchmark under the \emph{grounding-first} prompting setting.
``A-Acc'' denotes the answer accuracy.
``G-Acc'' denotes the grounding accuracy.
``Consist.'' denotes the answer-grounding consistency.
The best and second best results are marked in \textbf{bold} and \underline{underline}, receptively.
}
\vspace{-3mm}
\resizebox{1.0\textwidth}{!}{
\begin{tabular}{l|ccc|ccc|ccc|ccc}
\toprule[1.00pt]
\multicolumn{1}{c|}{\multirow{2}{*}{\textbf{Method}}} & 
\multicolumn{3}{c|}{\textbf{Attribute}} & 
\multicolumn{3}{c|}{\textbf{Judgement}} & 
\multicolumn{3}{c|}{\textbf{Object}}    & 
\multicolumn{3}{c}{\textbf{Average}}  \\
\multicolumn{1}{c|}{}   & 
\textbf{A-Acc.}  & \textbf{G-Acc} & \textbf{Consist.} & 
\textbf{A-Acc.}  & \textbf{G-Acc} & \textbf{Consist.} & 
\textbf{A-Acc.}  & \textbf{G-Acc} & \textbf{Consist.} & 
\textbf{A-Acc.}  & \textbf{G-Acc} & \textbf{Consist.} \\
\midrule
LLaVA-7B               \cite{llava-1.5} & 59.7     & 6.3        & 5.6  & 82.5 & 0.5     & 0.6         & 35.9 & 5.5     & 9.7        & 59.4 & 4.1     & 5.3         \\
LLaVA-13B              \cite{llava-1.5} & \underline{69.7}     & 11.5       & 11.7 & \underline{83.3} & 6.1     & 6.6         & \underline{51.9} & 14.6    & 17.2        & \underline{68.3} & 10.8    & 11.8        \\
LLaVA-OneVision-0.5B   \cite{llava_ov} & 0.0      & 0.0        & 0.0  & 57.3 & 0.0     & 0.0         & 3.0  & 0.0     & 0.0         & 20.1 & 0.0     & 0.0         \\
LLaVA-OneVision-7B     \cite{llava_ov} & 69.3     & 9.2        & 7.5  & 79.1 & 12.1     & 11.9         & 38.0 & 9.1     & 0.0         & 62.1 & 10.1     & 6.5         \\
LLaVA-OneVision-72B    \cite{llava_ov} & \bf{73.2}     & 17.2        & 14.0  & \bf{88.3} & 13.6    & 13.5        & \bf{56.8} & 17.9     & 17.7         & \bf{72.8} & 16.2     & 15.1         \\
InternVL2.5-4B         \cite{internvl2.5} & 27.2     & 36.2       & 22.8 & 68.0 & 17.5    & 18.1        & 14.6 & 40.7    & 14.5        & 36.6 & 31.5    & 18.5        \\
InternVL2.5-8B         \cite{internvl2.5} & 54.5     & 43.1       & 40.7 & 76.9 & 31.9    & 26.1        & 37.1 & 46.2    & 31.1        & 56.2 & 40.4    & 32.6        \\
InternVL2.5-38B        \cite{internvl2.5} & 58.2     & 71.2       & \bf{51.1} & 82.0 & 57.3    & \underline{49.5}        & 44.1 & \bf{51.1}    & \bf{44.2}        & 61.4 & 59.8    & \bf{48.3}        \\
InternVL2.5-78B        \cite{internvl2.5} & 63.4     & 41.8       & 33.1 & 79.1 & 42.7    & 41.0        & 34.0 & 58.4    & \underline{37.6}        & 58.9 & 47.6    & 37.2        \\
Qwen2.5-VL-3B-Instruct \cite{qwen2.5-VL}  & 26.1     & \underline{82.6}       & 25.1 & 37.4 & 66.0    & 26.8        & 23.4 & 55.6    & 31.3        & 29.0 & 68.1    & 27.7        \\
Qwen2.5-VL-7B-Instruct \cite{qwen2.5-VL}  & 48.8     & \underline{82.6}       & \underline{45.7} & 80.6 & \underline{72.8}    & \bf{62.4}        & 26.7 & \underline{62.3}    & 32.6        & 52.0 & \underline{72.6}    & \underline{46.9}        \\
Qwen2.5-VL-72B-Instruct\cite{qwen2.5-VL}  & 5.2      & \bf{85.0}       & 5.1 & 23.3 & \bf{73.8}    & 23.5        & 0.3  & \bf{67.5}    & 0.5       & 9.6  & \bf{75.4}    & 9.7        \\
\bottomrule[1.00pt]
\end{tabular}
}
\label{grounding_first}
\vspace{-3mm}
\end{table*}

\noindent \textbf{Dataset Construction.}  We follows a rigorous four-stage pipeline to build MM-GCoT, which is designed to ensure both reasoning complexity and grounding precision.
First, we align region descriptions with object annotations from Visual Genome \cite{vg}.
To ensure precise spatial correspondence, we employ IoU-based matching between these elements.
When an object has only one region whose IoU meets the threshold, we consider them to be a match.
Secondly, we use the matched objects as nodes to construct a spatial relationship graph based on spatial and semantic relations.
To generate multi-step reasoning chains, we iteratively sample the relationship paths from this graph.
Thirdly, we use structured templates to gather information of bounding box coordinates, object attributes, and contextual relationships.
Afterwards, we apply an LLM \cite{deepseekv3} to translate the templates into fluent natural language questions.
We implement a quality assurance process for both training and test samples.
For training data, we use LLM-based verification to check answer accuracy and grounding consistency.
Test samples are further validated manually to ensure the highest quality standards.


\noindent \textbf{Evaluation Metrics.} 
In terms of MM-GCoT evaluation, we mainly consider the evaluation for visual hallucination of MLLMs, since existing MLLMs are hard to inspire GCoT outputs without specific tuning. 
%
%
In this case, we focus on measuring the consistency of final answers only.
Following  previous VQA and Visual Grounding settings \cite{vqav2, RefCOCO}, we evaluate model performance using three key metrics: \emph{answer accuracy} (\textbf{A-Acc}), \emph{grounding accuracy} (\textbf{G-Acc}), and \emph{answer-grounding consistency} (\textbf{Consist.}).
A-Acc employs text matching between predicted and ground-truth answers, while G-Acc use Acc@0.5 as the metric \cite{RefCOCO}, requiring IoU $>$ 0.5 between predicted and ground-truth boxes. 
The answer-grounding consistency metric is defined by 
\begin{equation}
\begin{aligned}
Con. = \frac{|S_{ca,cb}|}{|S_{ca,cb}| + |S_{ca,wb}| + |S_{wa,cb}|},
\end{aligned}
\label{objective}
\end{equation}
where $S_{ca,cb}$, $S_{ca,wb}$ and $S_{wa,cb}$ denote sample sets with correct answer with correct box, correct answer with wrong box, and wrong answer with correct box, respectively. 
%
%

\noindent \textbf{Evaluation Prompt Settings for Existing MLLMs.} 
Most existing MLLMs are capable of visual grounding, but they are still hard to directly output answer with grounding information \cite{qwen2.5-VL, llava_ov}. 
In this case, we design two prompting strategy to inspire their grounded QA capability.
As illustrated in Fig.\ref{Fig_response}, the first solution is the \textbf{answer-first prompting}, which require the model to provide the answer first and then give the answer coordinates.
The second one is \textbf{grounding-first prompting}. 
It lets MLLMs first ground the potential answer, based on which answer the question. 
In terms of two settings, the answer-first one will be relatively easier, since the MLLM is to detect the answer instance. 
However, this setting can check whether the MLLM really see the answer cue in the image.
The other grounding-first setting is more challenging but also close to the target of our hallucination evaluation. 
It can examine whether the MLLM find the key visual element and answer the question based on it. 
All examined MLLMs will report the three metrics mentioned above on these two settings.

\section{Experiments}

\begin{table*}[t]
\caption{
Comparison between the original MLLMs and LLaVA GCoT.
``AF'' and ``GF'' refer to answer-first and grounding-first prompting settings, receptively.
``A-Acc'' denotes the answer accuracy.
``G-Acc'' denotes the grounding accuracy.
``Consist.'' denotes the answer-grounding consistency.
The best and second best results are marked in \textbf{bold} and \underline{underline}, receptively.
}
\vspace{-3mm}
\resizebox{1.0\textwidth}{!}{
\begin{tabular}{l|c|ccc|ccc|ccc|ccc}
\toprule[1.00pt]
\multicolumn{1}{c|}{\multirow{2}{*}{\textbf{Method}}} & 
\multicolumn{1}{c|}{\multirow{2}{*}{\makecell{\textbf{Prompt} \\ \textbf{Setting}}}} & 
\multicolumn{3}{c|}{\textbf{Attribute}} & 
\multicolumn{3}{c|}{\textbf{Judgement}} & 
\multicolumn{3}{c|}{\textbf{Object}}    & 
\multicolumn{3}{c}{\textbf{Average}}  \\
\multicolumn{1}{c|}{}   & 
\multicolumn{1}{c|}{}   & 
\textbf{A-Acc.}  & \textbf{G-Acc} & \textbf{Consist.} & 
\textbf{A-Acc.}  & \textbf{G-Acc} & \textbf{Consist.} & 
\textbf{A-Acc.}  & \textbf{G-Acc} & \textbf{Consist.} & 
\textbf{A-Acc.}  & \textbf{G-Acc} & \textbf{Consist.} \\
\midrule
Qwen2.5-VL-7B-Instruct \cite{qwen2.5-VL} & AF & \underline{73.6}     & \underline{72.5}       & \bf{59.8} & 87.9 & 56.3    & 51.5        & 57.8 & \bf{64.1}    & 59.1       & 73.1 & 64.3    & 56.8       \\
Qwen2.5-VL-7B-Instruct \cite{qwen2.5-VL} & GF & 48.8     & \bf{82.6}       & 45.7 & 80.6 & \bf{72.8}    & \underline{62.4}        & 26.7 & \underline{62.3}    & 32.6        & 52.0 & \bf{72.6}    & 46.9        \\
\midrule
LLaVA-7B               \cite{llava-1.5} & AF & 68.6 & 9.2 & 8.8 & 83.0 & 11.2 & 11.5 & 58.4 & 9.1 & 9.9 & 70.0 & 9.8 & 10.1        \\
LLaVA-7B               \cite{llava-1.5} & GF & 59.7     & 6.3        & 5.6  & 82.5 & 0.5     & 0.6         & 35.9 & 5.5     & 9.7        & 59.4 & 4.1     & 5.3         \\
\textbf{LLaVA-7B GCoT} & GCoT & 72.8     & 66.7       & 56.1 & \underline{88.3} & \underline{61.7}    & 56.9        & \underline{62.3} & 61.7    & \bf{61.3}        & \underline{74.5} & 63.3    & \underline{58.1}        \\
\midrule
LLaVA-13B              \cite{llava-1.5} & AF & 69.7 & 12.0 & 9.6 & 83.5   & 14.6 & 15.4 & 58.7   & 11.6        & 14.4 & 70.6 & 12.7 & 13.1        \\
LLaVA-13B              \cite{llava-1.5} & GF & 69.7     & 11.5       & 11.7 & 83.3 & 6.1     & 6.6         & 51.9 & 14.6    & 17.2        & 68.3 & 10.8    & 11.8        \\
\textbf{LLaVA-13B GCoT}& GCoT & \bf{74.7}     & 67.8       & \underline{58.0} & \bf{90.3} & \bf{72.8}    & \bf{68.0}        & \textbf{64.1} & 60.8    & \underline{59.3}        & \textbf{76.4} & \underline{67.1}    & \bf{61.8}    \\   
\bottomrule[1.00pt]
\end{tabular}
}
\label{table_comparison}
\end{table*}

\subsection{Implement Details}

We conduct a comprehensive evaluation of leading MLLMs through our GCoT dataset, selecting both pioneering architectures and state-of-the-art performers. 
The evaluated models include LLaVA (7B, 13B) \cite{llava-1.5} and LLaVA-OneVision (0.5B, 7B, 72B) \cite{llava_ov}, along with top-performing Qwen2.5-VL (3B, 7B, 72B) \cite{qwen2.5-VL} and InternVL2.5 (4B, 8B, 38B, 78B). 
To enable GCoT adaptation, we replace the original instruction-tuning samples with an equivalent number of examples from our GCoT dataset.
Then we train LLaVA GCoT from the scratch using this modified training set following the default settings of LLaVA.
%

\subsection{Quantitative Analysis}

\noindent \textbf{Evaluation on MM-GCoT dataset}
As shown in Tab.\ref{answer_first} and Tab.\ref{grounding_first}, we evaluate the performance of 12 state-of-the-art MLLMs on our MM-GCoT dataset using both answer-first and grounding-first strategies.
The evaluated models span diverse architectures and parameter scales, including LLaVA \cite{llava}, LLaVA-OneVision \cite{llava_ov}, Qwen2.5-VL \cite{qwen2.5-VL}, and InternVL2.5 \cite{internvl2.5}.
We can first observe that existing MLLMs exhibit severe visual hallucination issues.
For instance, while LLaVA-OneVision-72B achieves a 75.7\% average accuracy, it only maintains 11.1\% answer-grounding consistency under the answer-first prompt setting.
Notably, even the most advanced MLLM, \emph{e.g.} InternVL2.5-78B, still demonstrates limited accuracy-grounding consistency, achieving only 35.4\% on average.
Secondly, we observe that visual hallucination does not directly correlate with model scale.
The Qwen series shows marginal improvements in answer accuracy with only 1.3\% from 3B to 7B, and a mere 1.9\% when scaling to 72B.
Furthermore, regarding answer-grounding consistency, the Qwen2.5-VL-7B model exceeds its 72B by 18.2\%, reinforcing the observation that larger models do not guarantee better visual reasoning reliability.
Overall, these experiment results confirm the visual hallucination problem in current MLLMs.

\noindent \textbf{Comparison between Different Subsets.}
We then analyze the performance across different subsets, \emph{i.e.}, ``Attribute'', ``Judgement'' and ``Object'', as shown in Tab.\ref{answer_first} and Tab.\ref{grounding_first}.
From these tables, we can first observe that the ``Attribute'' subset demonstrates superior grounding accuracy
For instance, Qwen2.5-VL-7B achieves the grounding accuracy of 82.6\% under the grounding-first prompt setting.
For ``Attribute'' task, questions typically contain explicit spatial relationship descriptions, closely aligning with the grounding objectives used in MLLM training, thus yielding optimal grounding accuracy.
We can also observe that MLLMs achieve the highest answer accuracy in the ``Judgement'' task.
Specifically, the average answer accuracy achieves 86.1\% on average under the answer-first prompt setting.
``Judgement'' task, which primarily involve fact verification, present relatively straightforward challenges for MLLMs, resulting in higher answer accuracy.
Notebly, some MLLMs, \emph{e.g.} Qwen2.5-VL-3B, unable to strictly follow the instruction output under the grounding-first prompt setting.
It leads to the answer accuracy lower than 50\% on the ``Judgement'' task.
Another observation is that MLLMs show the strongest answer-grounding consistency in the ``Object'' task.
For example, the base model InternVL2.5-8B achieves 39.4\% answer-grounding consistency in the ``Object'' task.
In ``Object'' task, once MLLMs successfully identify the target, they can use the corresponding visual region to generate consistent responses, leading to superior consistency metrics.
Overall, these varied performance patterns across different task types demonstrate the multi-faceted nature of MLLM capabilities, highlighting how our GCoT dataset effectively evaluates different aspects of multimodal reasoning.

\begin{figure*}
\centering
\includegraphics[width=1.0\textwidth]{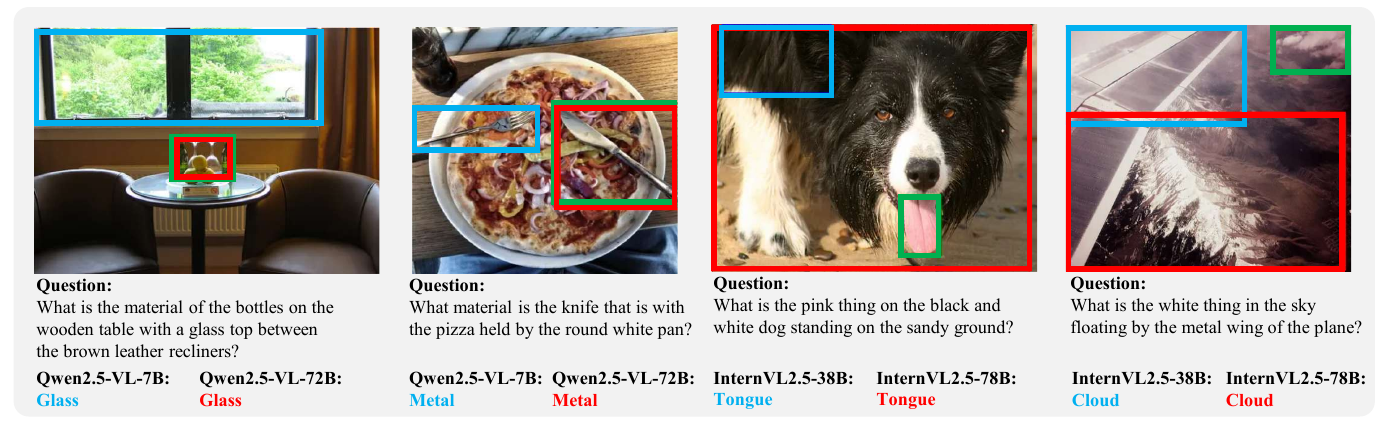}
\vspace{-7mm}
\caption{
Visualization of the results from Qwen2.5-VL and InternVL2.5 models of different parameter scales under \emph{answer-first} and \emph{grounding-first} promptings, respectively.
The \textcolor{myblue}{\textbf{blue}}, \textcolor{myred}{\textbf{red}} and \textcolor{mygreen}{\textbf{green}} bounding boxes represent predictions from base-scale models (Qwen2.5-VL-7B and InternVL2.5-38B), large-scale models (Qwen2.5-VL-72B and InternVL2.5-78B), and ground truth, respectively.
These visualizations show that super-large MLLMs exhibit poorer consistency in contrast.
}
\vspace{-3mm}
\label{fail}
\end{figure*}


\begin{figure*}
\centering
\includegraphics[width=1.0\textwidth]{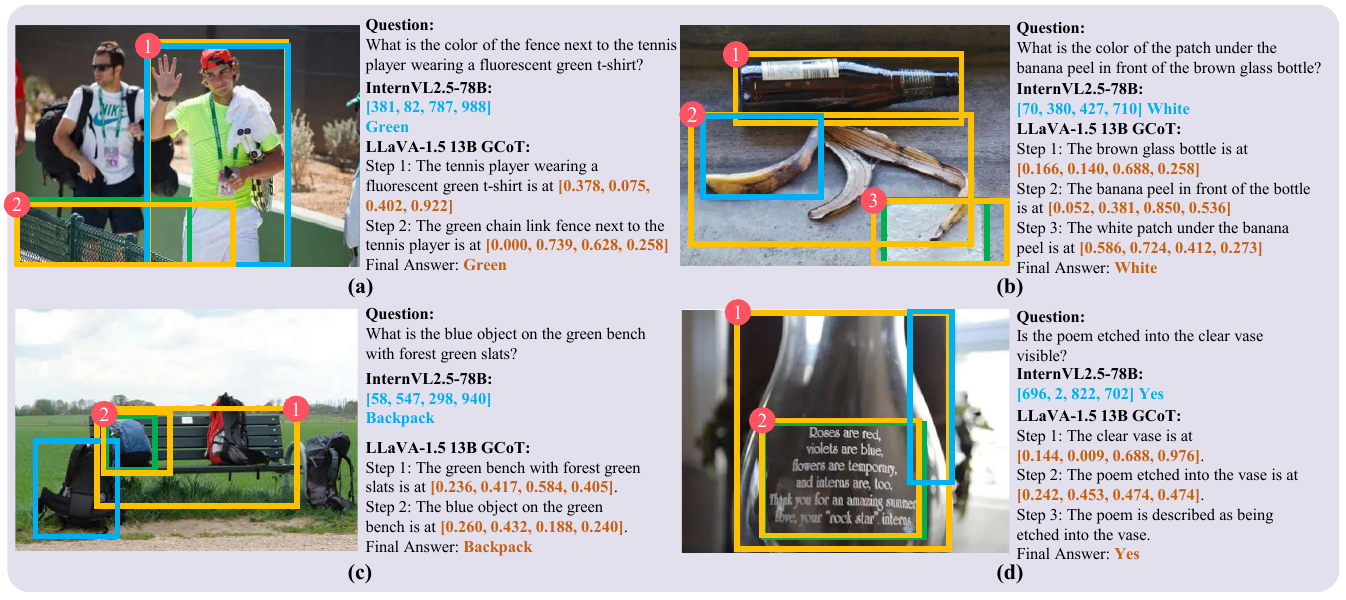}
\vspace{-7mm}
\caption{
Comparison between LLaVA-1.5 13B GCoT and InternVL2.5-78B. 
InternVL2.5-78B's responses and spatial predictions are visualized with \textcolor{myblue}{\textbf{blue}} bounding boxes. 
For LLaVA-1.5 13B GCoT, we display its answers, chain-of-thought reasoning process, and corresponding regions in \textcolor{myyellow}{\textbf{yellow}}. 
The ground truth regions are indicated by \textcolor{mygreen}{\textbf{green}} bounding boxes.
}
\label{Fig_comparision}
\vspace{-4mm}
\end{figure*}

\noindent \textbf{Comparison between Different Prompting Strategies.}
We further compare the experiment results between Tab.\ref{answer_first} and Tab.\ref{grounding_first}.
In the grounding-first setting, models are required to first identify relevant visual regions before answering questions.
We can first observe that MLLMs generally achieve higher accuracy in the answer-first approach.
For example, LLaVA-7B shows a 8.9\% higher accuracy under answer-first compared to the grounding-first prompt setting.
This discrepancy is particularly pronounced in smaller-scale models.
Notably, Qwen2.5-VL-3B exhibits a substantial accuracy gap of 45.4\% between the two approaches.
Furthermore, we can observe that while models show improved grounding performance in the grounding-first setting, their answer accuracy deteriorates.
For instance, Qwen2.5-VL-7B achieves a 7.9\% increase in Acc@0.5 for grounding compared to the answer-first approach, yet suffers an average accuracy drop of 21.2\% in answer generation.
These findings suggest that correctly question answering may not rely on explicit visual evidence.
Overall, these results indicate that MLLM does not consistently use visual and textual information during inference.

\noindent \textbf{Comparison between Different Model Scalings.}
We then compare the performance of the models from the same series but have different parameters.
We can observe that a model's visual grounding capability does not translate into related reasoning in responses.
For instance, while Qwen2.5-VL-72B-Instruct demonstrates superior grounding performance compared to Qwen2.5-VL-7B-Instruct, achieving a 2.4\% improvement on the ``Attribute'' task under the grounding-first prompt setting, the smaller model surprisingly exhibits better consistency scores with a 40.6\% margin.
Conversely, the MLLMs' reasoning ability does not automatically translate into enhanced grounding capability through textual information.
For example, while InternVL2.5-38B improves the answer accuracy by 5.5\% compared to its 8B counterpart, its grounding performance actually decreases by 0.3\% in the ``Object'' task.
Overall, these experimental results prove a gap between visual evidence and textual reasoning in current MLLMs, suggesting the need for more integrated training approaches.

\begin{figure}
\centering
\includegraphics[width=1.0\columnwidth]{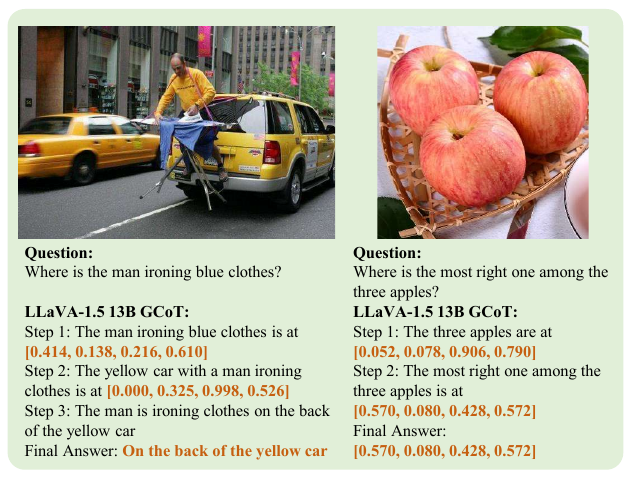}
\vspace{-7mm}
\caption{
Examples of LLaVA-1.5 13B GCoT on other tasks.
After training on our MM-GCoT, LLaVA-1.5 13B can transfer this capability to the tasks like common VQA and Visual Grounding.  
}
\label{general_GCoT}
\vspace{-4mm}
\end{figure}

\noindent \textbf{The effects of GCoT training}
We compare the performance of our LLaVA GCoT with state-of-the-art MLLMs in Tab.\ref{table_comparison}.
We can first observe that our GCoT training scheme significantly enhances the consistency and overall performance of MLLMs.
Specifically, LLaVA-7B GCoT achieves substantial improvements with a 4.5\% increase in accuracy and a remarkable 55.7\% boost in consistency compared to the original LLaVA-7B.
Furthermore, LLaVA-13B GCoT demonstrates competitive performance against Qwen2.5-VL-7B-Instruct, achieving comparable or better results across multiple metrics.
We can also observe that the improvement in consistency primarily stems from enhanced visual understanding rather than mere grounding capabilities.
For example, LLaVA-13B GCoT maintaining similar grounding performance (a slight increase of 2.2\%) while dramatically improving consistency by 61.2\%.
Overall, these results prove that GCoT effectively alleviate the limitation of current MLLMs in using visual information through explicit multimodal alignment.


\subsection{Qualitative Analysis}

\textbf{Comparison between Model Series.}
As shown in Fig.\ref{fail}, we visualize the results of the Qwen2.5-VL and InternVL2.5 models that perform under answer-first and grounding-first prompt setting, respectively.
We can first observe that Qwen2.5-VL-7B correctly predicts both the answer and corresponding regions in these two samples.
However, while Qwen2.5-VL-72B, correctly provides the answer, it identifies regions that share similar attributes with the target but are not actually relevant to the question.
This phenomenon demonstrates that merely scaling up model parameters and enhancing overall capabilities cannot eliminate the hallucinations that exist in MLLMs.
Under the grounding-first prompt setting, InternVL2.5 provides correct answers despite identifying mismatched regions.
The explicitly given error regions did not cause the model to answer the question incorrectly.
This suggests that even larger MLLMs do not necessarily rely on accurate visual evidence to generate their answers.
Overall, these experiment results reveal a gap between model scaling and visual grounding capabilities, highlighting the importance of explicit grounding mechanisms for reliable reasoning.

\noindent \textbf{Visualization of LLaVA GCoT.}
As shown in Fig.\ref{Fig_comparision}, we compare the performance of LLaVA-1.5 13B GCoT and InternVL2.5-78B through their responses and answer-grounding consistency.
We can first observe that LLaVA-1.5 13B GCoT effectively reduces hallucinations commonly observed in MLLMs.
For instance, in Fig.\ref{Fig_comparision}-(a) and (b), InternVL2.5-78B misidentifies objects with similar attributes and generates responses based on these incorrectly grounded regions.
In contrast, LLaVA-1.5 13B GCoT conducts step-by-step reasoning based on precise visual evidence, leading to accurate answers.
Furthermore, we observe that the inconsistency between answers and grounding becomes more pronounced in questions that lack direct object descriptions.
As demonstrated in Fig.\ref{Fig_comparision}-(c), even the state-of-the-art InternVL2.5-78B produces confused bounding box predictions when encountering similar objects.
However, LLaVA-1.5 13B GCoT successfully disambiguates between multiple potential targets through its chain-of-thought reasoning.
In some cases, even powerful MLLMs can exhibit severe hallucinations.
For example, in Fig.\ref{Fig_comparision}-(d), InternVL2.5-78B generates answers based on completely irrelevant regions.
In contrast, LLaVA-1.5 13B GCoT accurately identifies the target region while using only 16\% of the parameters.

Beyond the comparison, we visualize the reasoning and grounding results in the open-world QA and grounding tasks in Fig.\ref{general_GCoT}. 
From the figure, we can observe that LLaVA-1.5 13B GCoT generalizes well to tasks not included in our GCoT training datasets.
Specifically, it can effectively handle open-world questions while providing detailed visual evidence to support its reasoning.
Moreover, for referring expression comprehension tasks, the model generates answers with transparent and convincing reasoning processes.
Overall, these experimental results comprehensively demonstrate the effectiveness of GCoT in mitigating hallucinations in MLLMs while maintaining strong generalization capabilities.

\section{Conclusion}

In this paper, we propose a new learning task for MLLMs termed \emph{grounded chain-of-thought} (GCoT). 
GCoT aims to let MLLMs learn discrete visual reasoning steps with grounded information, thereby alleviating the visual hallucination issue. 
We also propose a new dataset for MLLMs called MM-GCoT, which can be used to train GCoT capability as well as evaluate the visual hallucination of existing experiments. 
Extensive experiments are conducted, and the experimental results not only yield notable findings about the hallucination of existing MLLMs, but also show the merits of GCoT towards trustworthy visual reasoning.

{
    \small
    \bibliographystyle{ieeenat_fullname}
    \bibliography{main}
}



%

\end{document}